\documentclass[runningheads]{llncs}
\usepackage{graphicx}
\begin{document}
\title{Integrating question answering and text-to-SQL in Portuguese}
%
%
\author{Marcos Menon José \inst{1}\orcidID{0000-0003-4663-4386} \thanks{Author to whom correspondence should be addressed.} \and
        Marcelo Archanjo José\inst{2}\orcidID{0000-0001-7153-0402} \and
        Denis Deratani Mauá\inst{3}\orcidID{0000-0003-2297-6349} \and
        Fábio Gagliardi Cozman\inst{1}\orcidID{0000-0003-4077-4935}}

\authorrunning{M. M. José et al.}

\institute{Escola Politécnica, Universidade de São Paulo, São Paulo, Brazil \\ \and Center for Artificial Intelligence (C4AI) \\
\and Instituto de Matemática e Estatística, Universidade de São Paulo, Brazil
\email{\{marcos.jose, marcelo.archanjo, denis.maua, fgcozman\}@usp.br}}
\maketitle              
\begin{abstract}
Deep learning transformers have drastically improved systems that
automatically answer questions in natural language. 
However, different questions demand different answering techniques; 
here we propose, build and validate an architecture that 
integrates different modules to answer two distinct kinds of queries.
Our architecture takes a free-form natural language text and classifies it to send it either to a Neural Question Answering Reasoner or a Natural Language parser to SQL. 
We implemented a complete system for the Portuguese language, using some of the main tools available for the language and translating training and testing datasets. 
Experiments show that our system selects the appropriate answering method with high accuracy (over 99\%), thus validating a modular question answering strategy.

\end{abstract}
%
%
%

\section{Introduction}

Question Answering (QA) has undergone a significant evolution over the past few years with the emergence of transformer neural networks  \cite{Vaswani2017}. Consider the Stanford Question Answering Dataset (SQuAD) task \cite{SQUAD20216}: state-of-the-art performance increased from about 70\% accuracy in 2018 to over 90\% in 2020~\cite{Zhang2020}, thus surpassing human performance. 

Despite their success, transformer-based neural models pre-trained on massive datasets   \cite{MARCO1018,SQUAD20216} still fail to answer satisfactorily certain types of questions.    
This is partly because many scalable approaches mostly repeat information in the corpus, and approaches that perform reasoning scale poorly \cite{Bender2021}.

In practice, it seems that different strategies better address different types of questions.
While factual and simple common-sense questions can be properly answered by neural question answering models with access to sizeable unstructured text corpora, questions that require complex reasoning or chaining of information are best answered by systems based on relational databases. 

To exploit the best of both approaches,  we propose here a new architecture that combines neural models trained with large, unstructured datasets, and neural models trained to respond to database queries states in natural language (text-to-SQL interfaces). 
Additionally, we target texts in the Portuguese language, as there has been little effort in developing question answering tools for this language. 

The proposed architecture takes natural-language questions and selects the appropriate answerer using a   text classifier. 
We experimented with two popular text classifiers: a naive Bayes classifier and a neural classifier based on the Portuguese pre-trained   transformer network BERTimbau (Portuguese BERT model) 
\cite{bertimbau}, the latter showing superior performance. 
Once a question is classified, it is fed  either into a neural reader-retriever model composed of a BM25 
\cite{Robertson2009} retriever and a PTT5 (Portuguese T5 model)
\cite{Carmo2021} reader, or into the text-to-SQL model mRAT-SQL (multilingual Relation-Aware Transformer SQL)
\cite{Jose2021} that uses a mT5
(multilingual T5 model)
\cite{xue-etal-2021-mt5} model to produce a SQL query
that is then run to generate an answer.

Our approach differs from previous efforts that combine techniques to QA. For example, in the landmark work on the Watson engine \cite{Ferrucci2011}, the question is feed to an ensemble of predictors and their confidences scores are then used to decide which answer to return (if any). The work of \cite{li2021dual} uses BM25 and T5, in which the first select both tables and passages related to the question, and then T5 has to decide and answer in SQL or text, thus using a single model to answer both types of questions.

We evaluated our approach on a new biomedical domain QA benchmark built by combining and translating to Portuguese three publicly available datasets: the MS MARCO dataset with factual questions \cite{MARCO1018}, and the text-to-SQL datasets Spider \cite{Yu2018} and MIMICSQL \cite{Wang2020}.
The use of a closed-domain (biomedicine) ensures that the classifier learns to differentiate between question types and not between different domains. 
All the relevant code, models and data are publicly available.\footnote{\url{https://github.com/C4AI/Integrating-Question-Answering-and-Text-to-SQL-in-Portuguese}}



\section{Background and Tools}

In this paper, we resort to text classifiers and question answering methods. 

{\bf Text classification} algorithms can be divided into classical approaches and deep neural networks. In the first case, in general, the text is transformed into a sparse vector and one of the conventional machine learning techniques, such as   Support Vector Machines, is used to classify it. The state-of-the-art deep learning approach uses pre-trained transformers networks since they allow the transfer learning, and the encoder layers can make good use of   context information in the input text.



\begin{figure}[t]
\includegraphics[width=\textwidth]{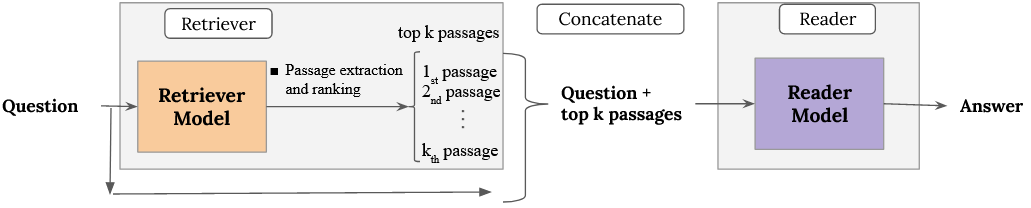}
\vspace*{-4ex}
\caption{Neural Question Answering Reasoner using the reader-retriever architecture.} 
\label{bm25ptt5}
\end{figure}

{\bf Question Answering systems that deal with factual questions} usually resort to two blocks: a retriever and a reader. The retriever    gathers   passages (small chunks of text) that are concatenated with the question;  the reader generates the answer (Fig.~\ref{bm25ptt5}). 
Some proposals use neural transformers both in the retriever and the reader \cite{RAG}, while others use keyword-based models for the retriever \cite{Chen17}.  

For the {\bf translation of textual questions to SQL}, known as NL2SQL, there are rule-based approaches, machine-learning approaches, and mixtures of both.
Unlike traditional QA, NL2SQL usually deals with context-specific questions and  idiosyncratic patterns. For example, the question ``\emph{How many appointments are there?}'' is meaningless without context, and must be grounded on the facts and schema of a specified database; e.g., a table named Appointment.



\section{Proposed Architecture}

\begin{figure}[t]
\includegraphics[width=\textwidth]{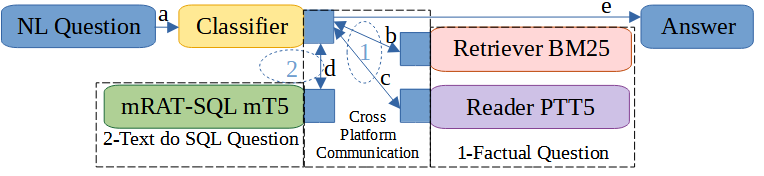}
\vspace*{-5ex}
\caption{Proposed architecture that deals both with factual and text-to-SQL questions.}
\label{Architecture}
\end{figure}

Fig.~\ref{Architecture} summarizes our proposal: 
a free-form natural language question is fed into a text classifier that outputs a classification of either 1-factual or 2-text-to-SQL, and processed accordingly.  Clearly, more classes can be added to the classifier to address other kinds of questions and other domains.
Factual questions are sent to a retriever (Fig.~\ref{Architecture}.b), which sends back passages related to the question. The question and the passages go to the reader (Fig.~\ref{Architecture}.c), which finally produces the answer (Fig.~\ref{Architecture}.e). 

We tested two text classifiers: a naive Bayes classifier with one-hot encoding of unigrams as features, and the BERTimbau language model, a version of the BERT model \cite{Devlin2018} that uses a bidirectional attention mechanism, pre-trained on a corpus of Brazilian Portuguese documents.

We used  BM25 \cite{Robertson2009}, a keyword-based technique, as  retriever, and PTT5 \cite{Carmo2021}, a Portuguese trained transformer-based encoder-decoder neural network model, as  reader. 
BM25 deals with sparse representation text document retrieval by ranking passages based on their similarity with the question text. This pipeline was based on the DEEPAGÉ system, an implementation by Cação et al.~\cite{cacao2021deepage}, which answered questions from the PAQ dataset \cite{lewis2021paq} about the environment translated into Portuguese.

Questions that require relational databases are sent to a dedicated NL2SQL module. One of the most successful techniques for this is the RAT-SQL+GAP \footnote{Relation-Aware Transformer SQL Generation-Augmented Pretraining.} system \cite{Shi2020} where, machine learning and a BART-large model are applied. We used a  multilingual version of that system, named mRAT-SQL \cite{Jose2021}, with a mT5-large model (Fig.~\ref{Architecture}.d) for Text-to-SQL, which sends back a query that answers the original question (Fig.~\ref{Architecture}.e).

A significant challenge is integrating these systems; each one has specific libraries, requirements and heavy computational demands. We developed a cross-platform communication scheme through the file system using a shared mount point to send and receive the questions and answers in different machines independent of their operating systems.

\section{Question Answering Datasets}

In order to build a system in Portuguese, it was necessary to use a dataset with questions and answers in that language. As most of the  resources are nowadays available in English, we applied automatic translation to existing datasets in English, thus translating questions using Google Translate and, in the case of MS MARCO, all the contexts and answers for training BM25+PTT5.  

The MS MARCO dataset \cite{MARCO1018} consists of questions that were anonymized from search queries of users of the Bing platform. One of its significant advantages is that it is based on actual questions that people ask in contrast to other artificially created datasets. In the end, human editors wrote down answers to each of the questions with the support of retrieved passages from texts on the internet.
To get questions from the closed domain of medicine, a partition created by Ying Xu et al.\ \cite{9206891} was used, which separated the question and answer pairs using   LDA topic modeling and clustering (they made the data accessible via API\footnote{The instructions to download can be found in the project github: \url{https://github.com/ibm-aur-nlp/domain-specific-QA}.}). Among the six available domains, the one of interest for this research was biomedicine, with 31620 question/answer pairs.
Table~\ref{tab1} shows some examples of translated natural language questions from the MS MARCO dataset.



%
To build a dataset for the classifier, 112 natural language questions were selected from   Spider \cite{Yu2018} as training dataset and 100 as   testing dataset.\footnote{Spider dataset is a popular resource that contains 200 databases with multiples tables under 138 domains: \url{https://yale-lily.github.io/spider}}
Those questions are related to 4 databases\footnote{hospital\_1 100 questions (test), protein\_institute 20 questions (train), medicine\_enzyme\_interaction 44 questions (train), scientist\_1 48 questions (train).} about  health and medical domains. 
Together with questions taken from Spider, we added a total of 10,000 questions taken from the MIMICSQL \cite{Wang2020} set, a dataset consisting of questions whose answers are SQL queries built automatically and reviewed by human editors. 
Table~\ref{tab1} shows more examples of natural language questions from   Spider   and from   MIMICSQL,\footnote{Text-to-SQL Generation for Question Answering on Electronic Medical Records Github: \url{https://github.com/wangpinggl/TREQS}.} both translated from English to Portuguese.

\begin{table}[t]
\centering
\caption{Questions Examples.}\label{tab1}
\vspace*{-2ex}
\begin{tabular}{| p{2 cm} | p{10 cm} |}
\hline
Dataset & Translated Question\\
\hline
MS MARCO & O que é fístula? \\
MS MARCO & De onde se ramifica a artéria descendente posterior?\\
Spider & Quais são os nomes dos pacientes que marcaram uma consulta?\\
Spider & Quais são os nomes dos cientistas designados para qualquer projeto?\\
MIMICSQL & quantos pacientes com menos de 45 anos?\\
MIMICSQL & encontre o número de pacientes únicos com diagnóstico de miopia.\\
\hline
\end{tabular}
\end{table}

To conclude this section, we offer a few comments on the datasets.
To build a model that correctly classifies each of the classes, the questions must have characteristics that distinguish them. Looking at   Table~\ref{tab1}, we see that some of the Text to SQL questions are quite specific, like: ``Quais são os nomes dos pacientes que marcaram uma consulta?'' (``What are names of patients who made an appointment?''). Instead, MS MARCO-derived questions like ``O que é fístula?" (``What is fistula?'') are much more factual and the answer must be accessed via a knowledge base or corpus.
Our dataset is a combination of two different types of datasets. One could argue that this introduces artifacts and trivializes the classification task. While we plan to evolve this research making the dataset ecologically valid \cite{McCallum2012}, we also argue that current QA datasets do share some of the characteristics of our fabricated dataset, with questions that are quite easy for humans to categorize into factual or SQL-based.


\section{Experiments}
 
Experiments were run in a number of machines:
AMD Ryzen 9 3950X 16-Core Processor, 64GB RAM, 2 GPUs NVidia GeForce RTX 3090 24GB running Ubuntu 20.04.2 LTS for both the PTT5 reader and mRAT-SQL; and AMD Ryzen 9 3950X 16-Core Processor, 32GB RAM, 2 GPUs NVidia GeForce RTX 3080 10GB running Ubuntu 20.04.2 LTS for the classifier and BM25.\footnote{We used well-known implementations of naive Bayes \cite{Zhang2004} and transformers. In particular, the tranformers HuggingFace library, at \url{https://huggingface.co/transformers/}, and also simpletransformers at \url{https://simpletransformers.ai/docs/installation/}.} 

\subsection{Classifier}
As the construction of each of the original datasets was different, it is possible to notice that there are some differences in the format of the questions. The biggest one is that in the MIMICSQL and Spider datasets, most questions came with punctuation (interrogation or period), while in MS MARCO this did not happen with the same frequency. Therefore, a cleanup was performed, taking away the final punctuation and some special characters so that the datasets would be more similar. This was important because the trained model did not learn to separate by unwanted information but to classify due to the content and type of question. 

The team decided to match the instance number between the two dataset classes for classifier training. So, there were 8000 training and 1000 validation MIMIC questions and 112 separate for Spider training (from protein\_institute, scientist\_1 and medicine\_enzyme: the ones related to biomedicine), we separated the first 9112 training questions from the MS MARCO dataset.

In order to obtain a correct evaluation of both models, Naive Bayes Classifier and BERTimbau,\footnote{Trained using 5 epochs, learning rate of 5e-5, batch size of 32 and maximum sequence length of 512.} and to compare them, cross-validation was performed in the database with 10 folds --- dividing the database into 10\% pieces and training with the other 90\% 10 different times. This allows a comparison between the models and to verify the standard deviation between the pieces of training and thus its generalizability. 

Once the best model is decided based on the cross-validation result, it will be tested by checking its performance separately for each dataset. All test questions from MS MARCO biomedical, 4743, all from test MIMIC, 1000, and 100 from the database Hospital\_1 from Spider were selected for this step.

\subsection{Question Answering Reasoner}
MS MARCO is made available with the questions gathered with their respective answers and the context of the answer. In order to simulate a real system, the questions were separated from their contexts, and the joining of all these documents formed the Knowledge Base (KB).

The BM25's function is to search inside this KB to help the reader answer the question. Thus, following the literature, the KB was broken into passages of up to 100 words (respecting punctuation), and when a question arrives, the model removes the K-passages closest to that question. The value of K chosen was 5 following the lighter model of the work by Cação et al.\ \cite{cacao2021deepage}. 

Finally, the PTT5 pre-trained neural network was trained \footnote{Trained using 25 epochs, learning rate of 2e-5, batch size of 32 and maximum sequence length of 512.} using the questions together with the respective 5 passages selected for the BM25 using the totality of the 22134 training questions of MS MARCO. After training, the model was tested with 4743 test questions and the Exact Match and Macro Average F1-Score metrics were verified based on an adaptation of the codes provided by the SQuAD dataset team.


\section{Results and Analyses}
The entire architecture has inference time that depends on the type of question it receives. While it takes an average of 7.4 seconds to answer a factual question, questions that are based on SQL tables take only 3.2 seconds on average. This was expected because the Question Answering Reasoner depends on two subsystems (BM25 and PTT5) while NL2SQL depends only on one (mRAT-SQL).

We start by analyzing the classifiers.
Both of them produced high F1-Scores\footnote{This is the standard F1-score for classification, not the Macro Average F1-Score.} in cross-validation with 10 different folds.
While the naive Bayes classifier is faster and lighter than BERTimbau, we decided to adopt the latter for the next steps as it got the best results in the validation (99.9\%), as shown in Table \ref{Validation}(top).

\begin{table}[t]
\caption{Top: results of the classifiers Naive Bayes and BERTimbau in a 10 fold cross validation in the training dataset. Bottom: results of the BERTimbau classifier in the different test datasets.}
\centering
\label{Validation}
\begin{tabular}{|c|c|}
\hline
\textbf{Classifier Algorithm} & \textbf{Validation F1-Score}   \\
\hline \hline
Naive Bayes Classifier & 98.2 $\pm$ 0.3 \%   \\
\hline
BERTimbau & 99.9 $\pm$ 0.01 \%   \\
\hline
\end{tabular}

\vspace{1ex}

\begin{tabular}{|c|c|c|}
\hline
\textbf{MS MARCO Test Accuracy} & \textbf{Spider Test Accuracy} & \textbf{MIMIC Test Accuracy}  \\
\hline \hline
 99.8 \%  &  98.0\% & 99.6\%  \\
\hline
\end{tabular}
\end{table}

In the testing datasets, BERTimbau  kept the best results from the cross-validation, Table \ref{Validation}(bottom). It is possible to infer that the results of Spider were slightly inferior because there were fewer training instances, but it was still excellent since the model managed to generalize from similar questions  MIMIC. 

These results demonstrate that questions in different classes are indeed quite different on average. Our proposed modular architecture does not significantly decrease the performance of the techniques that come in the following steps. 

As much as the BERTimbau based model has a close-to-perfect result, it is interesting to analyze which of the questions in the test set are wrongly classified. One of the questions in the MIMIC set is ``Como os Nephrocaps são administrados" (``\textit{How are Nephrocaps managed}''), which could very well be considered a factual question given that this information could come from a text. MS MARCO's question ``Qual é o propósito do pedido de Hektoen Enteric Agar'' (``\textit{What is the purpose of Hektoen's request Enteric Agar}'') was misclassified, probably because many of the SQL questions are about specific people like ``how many patients Doctor X has seen this month''.


Overall, results obtained with the MS MARCO Biomedical Dataset (translated into Portuguese) are consistent with the literature \cite{cacao2021deepage} with Macro Average F1-Score of 32.0\%. It is worth remembering that there is an inevitable variance depending on the dataset used, and, in the case of MS MARCO, there are responses that are longer than the PAQ dataset, which can increase the complexity. 

We should note that in our implementation, mRAT-SQL with mT5-large was fine-tuned with a multilingual dataset with four languages: English (original), Portuguese, Spanish, and French (these three translated versions using Google Translator). This extensive training dataset leads to the best results in Portuguese, even better than just training solely with the dataset in Portuguese~\cite{Jose2021}. 

In short, the architecture works as expected; we now present some examples that are related to an actual application.

When the model receives the question: ``O que causa dor nas costas'' (``\textit{What causes back pain}'') that is in the MS MARCO test group, it sends the question to the Question Answering Reasoner since the question seems factual and must look for the answer in texts. So, the final answer you get is: ``A dor nas costas é causada por uma queda ou levantamento pesado.'' (``\textit{Back pain is caused by a fall or heavy lifting.}''), since one of the retrieved texts is: ``A dor nas costas pode vir de repente e durar menos de seis semanas (aguda), que pode ser causada por uma queda ou levantamento pesado'' (``\textit{Back pain can come on suddenly and last for less than six weeks (acute), which can be caused by a fall or heavy lifting.}'').

As for the question ``Encontre o procedimento mais caro.'' (``\textit{Find the most expensive procedure.}''), the model correctly classifies it as an SQL question. Thus, the mRAT-SQL model returns the answer ``SELECT Procedures.Name FROM Procedures ORDER BY Procedures.Cost Asc LIMIT 1'' which performs all the queries correctly.


\section{Conclusion}

This paper proposed and validated a new Question Answering architecture that classifies natural language questions in Portuguese to feed them to appropriate systems, either a neural reader-retriever or a natural language parser to SQL. 
Experiments with real data demonstrated that a simple classifier can achieve near-perfect results in the closed biomedical domain. 
The possibility of integrating more subsystems is encouraging; while our original goal has been attained, there is still room for including other models and techniques. 

Since the dataset used merges different sources, it is possible that the results are overestimated.  Future work to mitigate this is to increase the number of sources and to perform data augmentation.


\section{Acknowledgments}

This work was partly supported by  Ita\'{u} Unibanco S.A.  through the \textit{Programa de Bolsas Ita\'{u}} (PBI) of the \textit{Centro de Ciência de Dados} da Universidade de São Paulo (C$^2$D-USP); by the Center for Artificial Intelligence (C4AI) through support from the S\~ao Paulo Research Foundation (FAPESP grant \#2019/07665-4) and from the IBM Corporation; by CNPq grants no.\ 312180/2018-7 and 304012/2019-0, and CAPES Finance Code 001.

\bibliographystyle{splncs04}
\bibliography{bibliography}

\end{document}